National Research Council Canada
Conseil national de recherches Canada
Institute for Information Technology
Institut de Technologie de l'information

# NRC·CNRC

*Increasing Evolvability Considered as a Large-scale Trend in Evolution**

P. Turney

Canada

NRC 43583

# Increasing Evolvability
# Considered as a Large-Scale Trend in Evolution

**Peter D. Turney**
Institute for Information Technology
National Research Council of Canada
Ottawa, Ontario, Canada
K1A 0R6
peter.turney@iit.nrc.ca

## Abstract

Evolvability is the capacity to evolve. This paper introduces a simple computational model of evolvability and demonstrates that, under certain conditions, evolvability can increase indefinitely, even when there is no direct selection for evolvability. The model shows that increasing evolvability implies an accelerating evolutionary pace. It is suggested that the conditions for indefinitely increasing evolvability are satisfied in biological and cultural evolution. We claim that increasing evolvability is a large-scale trend in evolution. This hypothesis leads to testable predictions about biological and cultural evolution.

## 1 INTRODUCTION

The idea that there is progress in evolution has been widely criticized (Gould, 1988, 1997). Progress implies that there is a large-scale trend and that the trend is good (Ayala, 1974, 1988). For example, it is commonly believed by the layperson that there is a large-scale trend in evolution towards increasing intelligence, and that this trend is good. McShea (1998) prefers to focus on the question of whether there are any large-scale trends, without regard to their value. He examines eight candidates for large-scale trends: entropy, energy intensiveness, evolutionary versatility, developmental depth, structural depth, adaptedness, size, and complexity (McShea, 1998). In this paper, we propose that evolvability (Dawkins, 1989, 1996) should be added to the list of candidates. We contend that evolvability is distinct from the eight candidates examined by McShea (1998). In particular, evolvability is not the same as adaptedness nor evolutionary versatility, although it might be argued that it subsumes these candidates.

It is difficult to define evolvability, beyond saying that it is the capacity to evolve. We suggest the following sufficient (but not necessary) condition for evolvability: If individuals *A* and *B* are equally fit but the fittest child of *A* is likely to be more fit than the fittest child of *B*, then *A* is more evolvable than *B*.[1] The point of this condition is that evolution does not directly select for evolvability, since (by hypothesis) *A* and *B* are equally fit. This is what makes evolvability especially interesting to evolutionary theorists. As Dawkins (1996) puts it, "This is not ordinary Darwinian selection but it is a kind of high-level analogy of Darwinian selection." There is no direct selection for evolvability, but there is nonetheless a large-scale trend towards increasing evolvability.

This might seem at first counter-intuitive. How can there be a large-scale trend towards increasing evolvability when evolvability does not enhance fitness? We present a simple computational model of evolvability and show that, in this model, evolvability increases, although it does not enhance fitness. This model is presented in the spirit of Hinton and Nowlan's (1987) influential computational model of the Baldwin effect. The idea of the model is to simplify to a bare minimum, in order to shed light on evolvability. In this model, part of the genome directly describes the phenome. The fitness of the phenome is completely determined by this part of the genome. The remainder of the genome influences evolvability. By construction, the rest of the genome can have no effect on fitness. However, in spite of this, our simulations show that there is a trend towards increasing evolvability. This model is similar to the model of Bedau and Seymour (1995).

After presenting the simple computational model of evolvability and the results of our simulations, we briefly discuss some of the possible mechanisms by which evolvability may increase in biology and in evolutionary computation. These mechanisms include kaleidoscopic development (Dawkins, 1989, 1996), modularity (Altenberg, 1994; Wagner and Altenberg, 1996; Simon, 1962; Turney, 1989), and the Baldwin effect (Hinton and Nowlan, 1987; Turney, 1996).

Finally, we discuss some of the testable (in principle) predictions of the claim that there is a large-scale trend towards increasing evolvability. The claim that evolvability is increasing implies that we should see an accelerating rate of evolution. We believe that analysis of biological and cultural evolution will support this prediction.

---

[1] We assume here that fitness is measured directly from the individual's phenotype. We do not mean fitness as measured by the long-term production of descendents.

## 2  A MODEL OF EVOLVABILITY

Our model of evolvability was implemented by modifying Whitley's (1989) GENITOR software.[2] GENITOR is a steady-state genetic algorithm (as opposed to a generational genetic algorithm) in which children are born one-at-a-time. A new child replaces the least fit member of the current population. We fixed the population size at 2000 individuals. The initial population of 2000 individuals was generated randomly. We then created a series of 5,000,000 children, using mutation, crossover, and selection.

In our model, an individual's genome is a string of 100 bits. The bits are arranged in pairs, where the first member of the pair (the odd bit) specifies the evolvability of the second member of the pair (the even bit). For a given genome, the corresponding phenome is the string of 50 bits that results when all the odd bits (the evolvability bits) are deleted from the genome. We measure the fitness of a phenome by comparing it with a target string of 50 bits. The bits in the target string are initially set randomly. Once every 8000 children, the target string is modified by randomly mutating 10% of its bits. We call this interval of 8000 children an *era*.

Parents are randomly selected, with a bias of 2.0 in favour of fitter individuals (see Whitley (1989) for details). We use single-point crossover, with the constraint that the crossover point cannot occur within a pair; it must occur between two pairs.[3] Mutation is randomly applied to 50% of the children, after crossover. If a child is chosen for mutation, then the odd bits (evolvability bits) are randomly mutated with a probability of 0.0001. The even bits (phenome bits) are mutated with a probability of 0.01, but only if the corresponding evolvability bit is set to 1. When the evolvability bit is set to 0, the phenome bit cannot mutate.

The fitness of a phenome is calculated by counting the number of bits that match with the 50 bit target string. This number is converted to a frequency by dividing it by 50. The fitness is then calculated by raising the frequency to the power of 10. Thus a perfect match yields a fitness of 1.0. A match of 49 bits yields a fitness of 0.817. This fitness function gives a strong incentive to match all the target bits.

This model has been designed so that the evolvability bits can have no direct impact on the fitness. By design, there can be no direct selection for evolvability. However, since the target shifts once each era, there is an advantage to being able to adapt to the shifting target.

Figure 1 shows the result of a typical run of this model. We measure evolvability by the percentage of odd bits in the genome that have the value 1. We measure the fitness at the start and end of an era. The measurements of evolvability and fitness are averages over the whole population of 2000 individuals. The population averages are calculated once every 500 children. These averages are then combined for each group of 100,000 children. Thus each curve begins at the 100,000th child, at which point we have 200 observations of the population's average evolvability (100,000 divided by 500) and 12 observations of the population's average fitness at the start and end of an era (100,000 divided by 8000).

Since the initial population is random, the expected average evolvability of the population at the time of the first child is 50%. This rapidly climbs to about 60% by the time the 100,000th child is born. When we reach the 5,000,000th child, the evolvability is about 95%. At this point, it appears to have achieved an equilibrium point with the mutation rate for evolvability bits (0.0001). If we run the simulation until the 10,000,000th child, the evolvability is still about 95%.

As evolvability increases, the average fitness increases, at both the start and end of an era. We define the fitness improvement during an era as the average fitness at the end of an era minus the average fitness at the start of an era. The fitness improvement is an indicator of the pace of evolution. The second plot in Figure 1 shows that the fitness improvement is increasing. Thus the second plot shows that evolution is accelerating.

## 3  DISCUSSION

In our model, evolvability cannot increase without limit, because the genome length is fixed at 100 bits. However, we believe that this limitation could be overcome (at the cost of increased complexity in the model) by allowing the genome length to vary. If the genome length can grow without bound, evolvability can increase without limit.

Although the model is relatively simple, we do not fully understand its behaviour. There are many parameters in this model (the population size, the phenome bit mutation rate, the evolvability bit mutation rate, the length of an era, the fitness function, etc.). We would like to be able to specify the parameter settings for which evolvability will increase until it reaches the limits set by the genome length and the evolvability bit mutation rate. It will require many experiments to explore the parameter space of this model.

One preliminary observation is that the model is sensitive to the relative values of the mutation rates for the evolvability bits (0.0001) and the phenome bits (0.01). The evolvability bits are not directly selected, unlike the phenome bits, which means that the evolvability bits are more vulnerable to disruption by mutation. For evolvability to increase, the evolvability bits must persist long enough for their value to become apparent. This observation may be helpful for practical applications of evolutionary computation. In practical applications, to ensure evolvability, it may be necessary to design evolutionary systems with a built-in mechanism for protecting parts of the genome from the disruptive effects of high mutation rates. We predict that this also applies to biological evolution. Of course, our model strictly separates the genes for evolvability from the genes for the phenome, for illustrative purposes. In cases where there is not such a strong separation, there will be less need for different mutation rates.

---

[2] GENITOR is written in C. The source code is available on the Internet at ftp://ftp.cs.colostate.edu/pub/GENITOR.tar.

[3] In other words, the bit to the left of the crossover point is even and the bit to the right is odd. We have not empirically tested whether this constraint has any effect. It seemed sensible to prevent crossover from breaking up pairs.

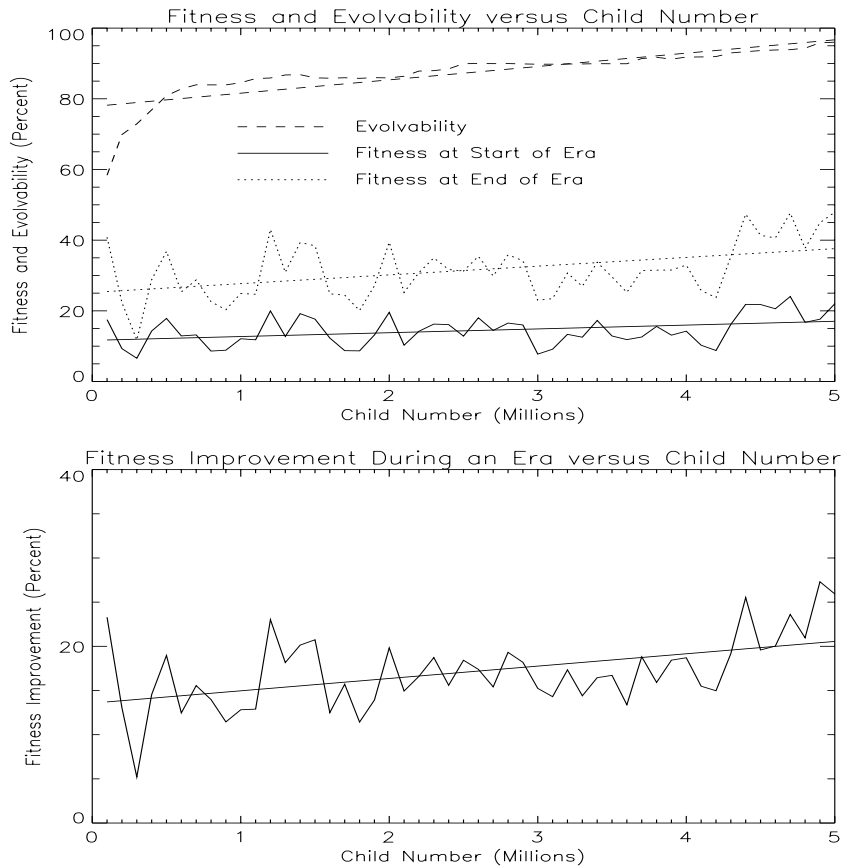

Figure 1: The plot at the top shows increasing trends in evolvability, fitness at the start of an era, and fitness at the end of an era. The plot at the bottom shows that the spread between the fitness at the start and end of an era is also increasing. This implies that the pace of evolution is increasing. All of the curves have been fitted with straight lines, using linear regression, to make it easier to see the trends.

Another preliminary observation is that the length of the era is important. If the era is too long, there is an advantage to setting the evolvability bits to 0, in order to lock-in the phenome bits that match the target, to protect them from mutation. If an era is too short, there is not enough time for the advantage of evolvability to become apparent. We predict that this also applies to biological evolution. For evolvability to increase, environmental change must occur within certain bounds. If there is too little change, there is no advantage to evolvability. If there is too much change, evolution cannot move fast enough to track the changes.

The model presented here is highly abstract. It does not consider the mechanisms necessary to achieve evolvability in biological evolution or applied evolutionary computing. However, we believe that the mechanisms for evolvability that have been discussed in the literature are consistent with this abstract model. Kaleidoscopic development (Dawkins, 1989, 1996), modularity (Altenberg, 1994; Wagner and Altenberg, 1996; Simon, 1962; Turney, 1989), and the Baldwin effect (Hinton and Nowlan, 1987; Turney, 1996) can all be viewed as mechanisms that enable a certain aspect of the phenotype to change by genetic mutation, where such mutation would be detrimental without these mechanisms. In effect, without evolvability, certain mutations are forbidden; certain evolutionary paths are closed. We model this very simply and abstractly, by locking bits so that they cannot mutate.

The model was partly inspired by Hinton and Nowlan's (1987) model of the Baldwin effect. Hinton and Nowlan (1987) demonstrated that phenotypic plasticity (specifically, lifetime learning by individuals) can enable an evolutionary system to find optima that would be very difficult to find without plasticity. Our model is also related to Anderson's (1995) model of the Baldwin effect. In Anderson's (1995) model, the effect of learning is represented as an increase in the variance of selection. He derives equations and equilibrium conditions for a population of learning individuals under fixed and variable environmental selection. Other related work is the model of Bedau and Seymour (1995). In Bedau and Seymour's (1995) model, mutation rates are allowed to adapt to the demands of the environment. They find that mutation rates adapt to an optimal level, which depends on the evolutionary demands of the environment.

One of the most interesting implications of the model is that increasing evolvability results in accelerating evolution. We believe that the pace of biological and cultural evolution is accelerating. None of the eight candidates for large-scale trends (entropy, energy intensiveness, evolutionary versatility, developmental depth, structural depth, adaptedness, size, and complexity) that were examined by McShea (1998) imply that the pace of evolution should accelerate.[4]

It is difficult to objectively verify the claim that the pace of evolution is accelerating. We can look at the historical frequency of innovations, but the analysis is complicated by several factors. One confounding factor is that our record of the recent past (both the fossil record of biological evolution and the record of cultural evolution) is superior to our record of the distant past, which may give the illusion that there are more innovations in the recent past than the distant past. For cultural evolution, another confounding factor is population growth. We may expect more innovations in recent cultural history simply because there are more innovators. A third factor is difficulty of counting innovations. We need to set some kind of objective threshold on the importance of the innovations.

We suggest some tests that avoid these objections. In paleobiology, we predict that the fossil record will show (1) an accelerating rate of spread of life into various previously sterile territories, (2) decreasing recovery time from major catastrophes (e.g., mass extinction events, ice ages, meteorites), and (3) a decrease in the average lifetimes of species, as they are out-competed by more recent species at any accelerating rate (i.e., acceleration of obsolescence). These three tests do not involve counting the frequency of innovations, which makes them relatively objective.

We believe that accelerating evolution due to increasing evolvability is also manifest in cultural evolution. For example, consider the decrease in the average lifetime of new technology (acceleration of obsolescence). However, technologies do not form species, which makes it difficult to objectively measure the lifetime of a technology. We also need to correct for population growth, perhaps by comparing the estimated pace of cultural evolution to the estimated pace of population growth.

Kurzweil (1999) points out that the rate of evolution of computing machinery is accelerating. For many years, the speed of computers has roughly doubled every two years. More recently, the doubling time has reduced to 18 months. We predict that many other technologies will display this same pattern of evolution. For example, we could look at the information density in magnetic recording media. The problem with specific measures of this type (e.g., computation speed, information density) is that they may be inappropriate when there is a major revolution in technology (e.g., quantum computers, holographic memory).

---

[4] Unfortunately there is not enough space here to properly defend this claim. Note that McShea (1998) points out that several of the candidates imply that the average lifetime of species should be increasing, whereas we predict that the average lifetime of species should be *decreasing*, due to accelerated obsolescence.

**Acknowledgements**

Thanks to Russell Anderson and the anonymous referees for helpful comments.